\documentclass[journal]{IEEEtran}

\IEEEoverridecommandlockouts  
\usepackage[utf8]{inputenc}
\usepackage[english]{babel}
\usepackage{graphicx}
\usepackage{amsmath, amsfonts, amssymb}
\usepackage{algorithm, algpseudocode}
\usepackage{mathrsfs}
\usepackage{lineno}
\usepackage{rotating}
\usepackage{xcolor}
\usepackage{csquotes}
\usepackage[inline]{enumitem}
\usepackage{wrapfig}
\usepackage{booktabs}
\usepackage{multirow}
\usepackage{dblfloatfix}
\RequirePackage{array}
\RequirePackage{arydshln}
\usepackage[font=footnotesize]{subcaption}
\usepackage{placeins}
\usepackage{url}

\title{\LARGE \bf
An Adaptive Human Driver Model for Realistic Race Car Simulations
}

\author{Stefan Löckel$^{1}$, Siwei Ju$^{2}$, Maximilian Schaller$^{3}$, Peter van Vliet$^{4}$, and Jan Peters$^{5}$ 
\thanks{*This work was supported by Dr. Ing. h.c. F. Porsche AG}
\thanks{$^{1}$Stefan Löckel is with Computer Science Department,
        Technische Universität Darmstadt, 64289 Darmstadt, Germany and with Dr. Ing. h.c. F. Porsche AG, 71287 Weissach, Germany
        {\tt\small loeckel@ias.tu-darmstadt.de}}%
\thanks{$^{2}$Siwei Ju is with Computer Science Department,
        Technische Universität Darmstadt, 64289 Darmstadt, Germany and with Dr. Ing. h.c. F. Porsche AG, 71287 Weissach, Germany
        {\tt\small siwei.ju@porsche.de}}%
\thanks{$^{3}$Maximilian Schaller is with ETH Zürich, 8092 Zürich, Switzerland
        {\tt\small mschaller@ethz.ch}}%
\thanks{$^{4}$Peter van Vliet is with Dr. Ing. h.c. F. Porsche AG, 71287 Weissach, Germany
        {\tt\small peter.van\_vliet@porsche.de}}%
\thanks{$^{5}$Jan Peters is with Computer Science Department,
        Technische Universität Darmstadt, 64289 Darmstadt, Germany
       {\tt\small peters@ias.tu-darmstadt.de}}%
}

\usepackage[absolute,showboxes]{textpos}

\TPMargin{2.5pt}

\newcommand{\IEEEstatement}{
    \begin{textblock}{0.84}(0.08,0.93)    
         \centering
         \scriptsize
         This work has been submitted to the IEEE for possible publication. Copyright may be transferred without notice, after which this version may no longer be accessible.
    \end{textblock}
}
\begin{document}
	\maketitle
    \IEEEstatement 
	\begin{abstract}
Engineering a high-performance race car requires a direct consideration of the human driver using real-world tests or Human-Driver-in-the-Loop simulations.
Apart from that, offline simulations with human-like race driver models could make this vehicle development process more effective and efficient but are hard to obtain due to various challenges.
With this work, we intend to provide a better understanding of race driver behavior and introduce an adaptive human race driver model based on imitation learning.
Using existing findings and an interview with a professional race engineer, we identify fundamental adaptation mechanisms and how drivers learn to optimize lap time on a new track.
Subsequently, we use these insights to develop generalization and adaptation techniques for a recently presented probabilistic driver modeling approach and evaluate it using data from professional race drivers and a state-of-the-art race car simulator.
We show that our framework can create realistic driving line distributions on unseen race tracks with almost human-like performance.
Moreover, our driver model optimizes its driving lap by lap, correcting driving errors from previous laps while achieving faster lap times.
This work contributes to a better understanding and modeling of the human driver, aiming to expedite simulation methods in the modern vehicle development process and potentially supporting automated driving and racing technologies.
\end{abstract}
	\section{Introduction}
\begin{figure}
	\centering
	\includegraphics[width=\columnwidth]{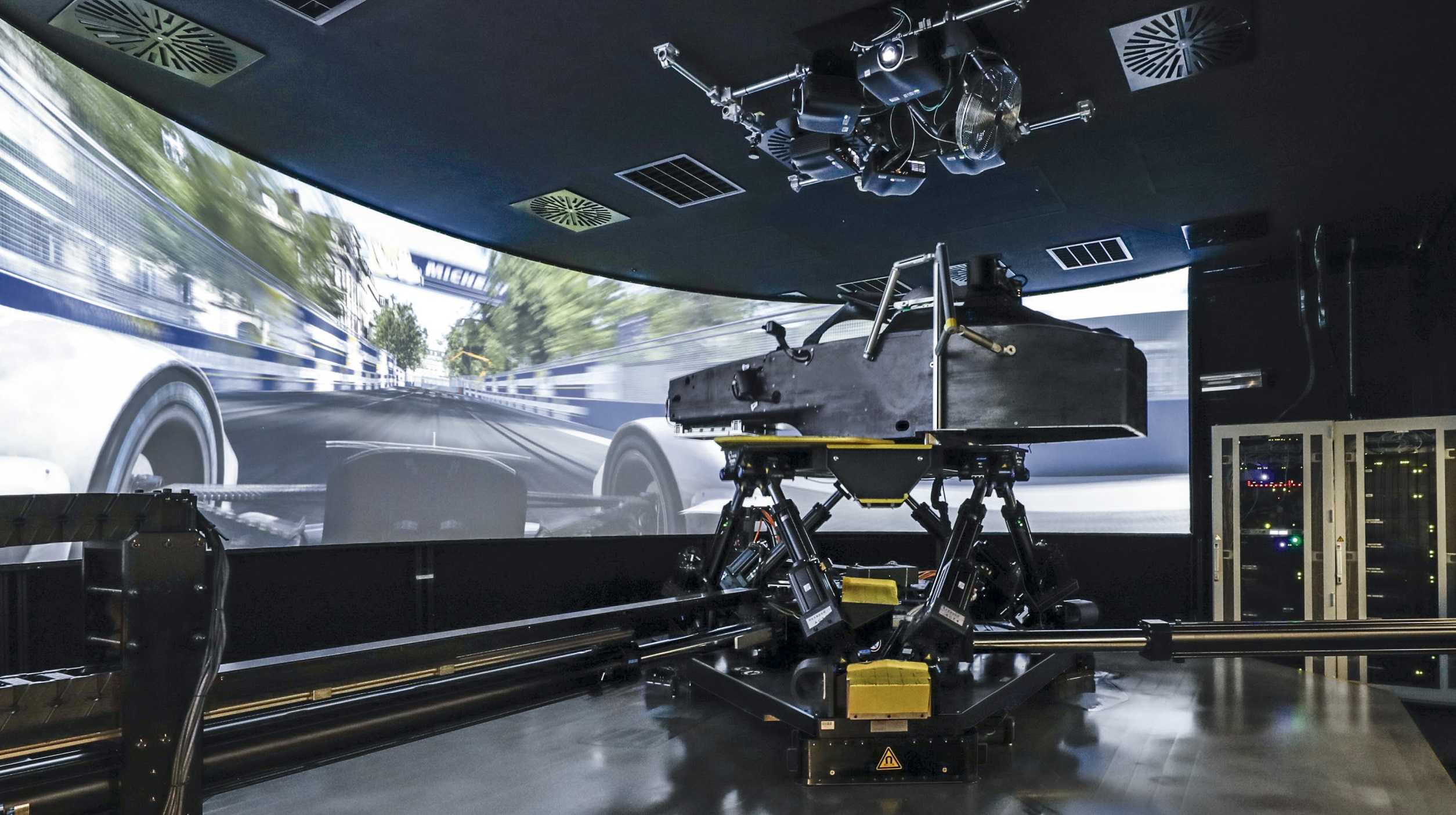}
	\caption{Race car simulator at Porsche Motorsport \cite{porsche:2019:SIM}:
	A realistic visualization, a vehicle cockpit mounted on an actuated platform, and a complex, in-house developed vehicle dynamics simulation facilitate rapid testing of new vehicle configurations with the human driver in the loop.
	The adaptive human driver model is developed using demonstration data from professional race drivers which was generated in this simulator.
	Subsequently, the driver model is evaluated using the same simulation environment, intending to support the future vehicle development process.}
	\label{fig:simulator}
\end{figure}

Throughout motorsports' more than 125 years long history, the fundamental goal of all participants did not change: reaching the best racing performance among competitors, which ultimately requires engineering a race car that fits its driver well.
In fact, Milliken \& Milliken already stated in 1995 that \enquote{It is the dynamic behavior of the combination of high-tech machines and infinitely complex human beings that makes the sport so intriguing for participants and spectators alike} \cite{milliken:1995:RCVD}.
Hence, for modern vehicle development in professional motorsports, a good understanding and modeling of the human driver are crucial to further improve the performance of the driver-vehicle-system.
At the same time, the human decision-making process during racing is extremely complex and thus difficult to model, as:
\begin{enumerate}
	\item many influencing factors exist,
	\item vehicle dynamics are highly nonlinear and race cars are usually driven at the handling limits, and thus difficult to control,
	\item each driver exhibits an individual driving style,
	\item the generalization and adaptation mechanisms are complex and difficult to incorporate in a driver model.
\end{enumerate}
While challenges 1-3 have been successfully addressed in recent research \cite{loeckel:2020:PFIH, loeckel:2021:IMRD}, the problem of understanding and imitating the human adaptation process remains unsolved.
With our work, we intend to identify and understand the most important adaptation and learning techniques mastered by professional race drivers, contribute to the modeling of driver behavior by developing two methods to mimic this behavior, and evaluate the proposed methodology within a realistic Human-Driver-in-the-Loop (HDiL) simulator as shown in Figure \ref{fig:simulator}.
\par
A well-fitting driver model is not only useful for advancing autonomous driving and racing technology but also to support the modern vehicle development process.
Such a model could considerably extend and improve full vehicle simulations, ultimately enhance the resulting vehicle performance and development efficiency, while being much less expensive compared to HDiL simulations.
\subsection{Problem Statement and Notation}
In order to model human race driver behavior we aim to learn a human-like control policy $\pi^M$ which maps the current overall state $\mathbf{x}$, including vehicle state and situation on track, to the vehicle control inputs $\mathbf{a} = \left[\delta~g~b\right]$ composed of steering wheel angle $\delta$, throttle pedal position $g$ and brake pedal actuation $b$.   
This policy should be able to robustly maneuver a race car at the handling limits while being similar to the unknown internal driving policy $\pi^E$ of human experts.
At the same time, this expert policy is non-deterministic due to natural human imprecision and intentional adaptation, and able to generalize to new situations as, for example, new race tracks.
In this work, we aim to approach the problem of modeling this behavior by:
\begin{enumerate}[label=(\alph*)]
	\item identifying and understanding the most important adaptation and learning mechanisms through related work and an expert interview with a professional race engineer,
	\item using these findings to considerably extend a data-based driver modeling approach, and
	\item evaluating the developed methods using data from professional race drivers and a state-of-the-art motorsport simulation environment.
\end{enumerate}
Consequently, the resulting driver-specific control policy $\pi^M$ should be able to generalize to unseen tracks and exhibit certain adaptation characteristics as the human driver.
\subsection{Related Work}
As vehicle dynamics are well understood nowadays, different approaches with varying complexities are available to model the physics of a car in different driving situations \cite{milliken:1995:RCVD,schramm:2010:MSDK,perantoni:2014:OCFO,siegler:2002:LTSR,baruh:2014:AD}.
Such vehicle models can be used to predict the driving behavior in standard maneuvers or to estimate the vehicle performance on a particular race track using lap time simulation approaches \cite{siegler:2002:LTSR,siegler:2000:LTSC,gadola:1996:TLTS,kelly:2008:LTST}.
However, individual human driver behavior, being an important component of the vehicle-driver-entity, is often not sufficiently considered by these methods.
This fact encourages motorsport teams to utilize HDiL simulation approaches, where the real driver operates the vehicle within a realistic simulator environment, facilitating faster prototyping and more realistic predictions of the true vehicle performance \cite{fritzsche:2017:SBDP}.
\par
A variety of related work describes car racing from the driver's perspective, analyzes racing techniques, driving lines, and the complex decision processes in greater detail, and contributes to a better understanding of the human driver in general \cite{bentley:2011:USSC,bentley:1998:SSPR,krumm:2015:DEAS,kegelman:2018:LPRC}.
Nevertheless, the task of modeling this behavior remains highly challenging.
A number of approaches for building a virtual driver for different use cases mainly rely on conventional control architectures with a limited number of parameters \cite{hess:1990:CTMD}, on model predictive control \cite{novi:2019:CSFA}, and on exact linearization \cite{koenig:2009:VTQG}.
Some recently developed methods utilize machine learning techniques to imitate human drivers based on demonstration data:
using supervised learning, random forests were trained to predict car control inputs from basic vehicle states \cite{cichosz:2014:ILCD} and it was shown that a feedforward neural network is able to track a driving line generated by a human \cite{wei:2013:MHDB}.
Furthermore, methods based on reinforcement learning like the Generative Adversarial Imitation Learning (GAIL) framework \cite{ho:2016:GAIL} were utilized to mimic drivers in a highway driving scenario \cite{kuefler:2017:IDBG}, and were extended to imitate human behavior in a short-term race driving setting based on visual features \cite{li:2017:IIIL}.
Besides that, research on personalized driver modeling is available that targets specific human individuals \cite{deng:2020:PMDS,schnelle:2017:DSMP,shi:2015:EDSN}.
However, these models do not consider human adaptability and are partially limited to only model steering as a car control input.
\par
The \textit{Probabilistic Modeling of Driver Behavior} (ProMoD) framework was demonstrated to be capable of completing full laps with a competitive performance by mimicking professional race drivers in a realistic HDiL simulation environment \cite{loeckel:2020:PFIH,loeckel:2021:IMRD}.
The data-based and modular approach learns distributions of driving lines represented by Probabilistic Movement Primitives (ProMPs) \cite{paraschos:2013:PMP,paraschos:2018:UPMP} and trains a recurrent neural network on human race driver data in a supervised fashion.
Related to this, there seems to be a shift from linear and time-invariant models of human manual control to nonlinear and time-varying approaches that are apparent in current research trends \cite{mulder:2017:MCCS}.
In particular, adaptation over time is identified as a key aspect of human behavior that should and can be modeled by moving towards time-varying models.
While the ProMoD framework is shown to work well in many situations, it is still lacking the functionality of a time-varying model, i.e., the ability to learn driving on unknown tracks and to adapt and learn from gathered experience from driven laps.
As such learning and adaptation aspects play fundamental roles in competitive motorsports, any robust and accurate driver modeling approach should be able to reflect them.
\par
Human adaptation behavior w.r.t. adaption times for changing road types in a driving simulator is analyzed, yet not modeled in the work of \cite{ronen:2013:APDS}.
Past research on modeling driver adaptation to sudden changes of the vehicle dynamics takes into account limb impedance modulation and updating of the driver internal representation of the vehicle dynamics \cite{deborne:2012:MDAC}.
However, this related work focuses exclusively on the lateral dynamics with a first-principles approach without a superordinate objective such as lap time.
\par
With our work, we considerably modify and extend ProMoD to model human driving adaptation - to the best of our knowledge, for the first time in the racing context.
Due to the modular architecture, the driving policy is adapted in a transparent manner. 
We contribute to a better understanding of human race driver behavior, considerably enhance the quality of a modern driver modeling approach, and aim to pave the way for more accurate vehicle simulations and, potentially, future autonomous racing.
	\section{Methodology}
As a proper understanding of the human race driver is fundamental for modeling its learning techniques, we ground our methodology on key insights from literature, supplemented by findings of an expert interview with a professional race engineer\footnote{A race engineer works at the interface between the driver and the vehicle, trying to help the driver work with the vehicle and to find a vehicle setup tailored to the driver's needs.} for LMP1\footnote{Le-Mans-Prototypes represent a top class of race cars used in different endurance racing series with races lasting up to 24 hours.} race cars.
The adaptation principles identified in Section \ref{sec:expert_interview} are followed by a short summary of the recently presented ProMoD driver modeling framework in Section \ref{sec:promod_original}.
In Section \ref{sec:trackgen} we present a novel way to generalize the driver model to new tracks.
Finally, Section \ref{sec:featureadapt} introduces a new method to optimize driving similar to a real race driver based on experience from previous laps.
\subsection{Adaptation Principles}
\label{sec:expert_interview}
Race drivers constantly pursue better racing performance in the presence of new tracks and modified vehicle setups.
In this section, we aim to understand the most important principles for the adaptation behavior of race drivers.
We gather the following key insights from literature, extended with an expert interview\footnote{Findings from the expert interview are marked with this footnote. A summary of the interview is given in Appendix \ref{app:expertinterview}.} of a professional race engineer in Appendix \ref{app:expertinterview}:
\paragraph*{Objective: (delta) lap time} Drivers aim to drive as fast as possible and minimize the lap time in order to win races \cite{braghin:2008:RDM}.
Hence, drivers tend to pay attention to their delta lap time, i.e. relating the current lap to the previous or the best lap.$^\text{3}$
Any modifications to the vehicle setup or environmental influences are handled as disturbances by adapting the control policy.$^\text{3}$
\paragraph*{Risk awareness} Race drivers are particularly risk-aware and constantly test for the vehicle limits \cite{kegelman:2018:LPRC}.
Furthermore, they aim to optimize performance by starting from a safe region and improving their driving incrementally.$^\text{3}$
\paragraph*{Hierarchy} Brake points and speed profile are related hierarchically in a sense that, starting from anchoring and shifting the brake points, the corner-entry speed follows as a consequence and influences the performances of the entire corner \cite{bentley:1998:SSPR,krumm:2015:DEAS}.
Finally, the driving line follows from the speed profile and the driver tries to control these three aspects in the same hierarchical order.$^\text{3}$
\paragraph*{Initialization - Driving on new tracks} When starting on a new track, drivers tend to compare all new situations and corners to their experience from other tracks \cite{bentley:1998:SSPR,krumm:2015:DEAS}.
This information is used to get an initial guess of reasonable brake points and driving lines, which is subsequently refined.$^\text{3}$
The initialization of brake points begins already before starting to drive, while the speed profile and driving line is initialized during the first few laps.$^\text{3}$ After initialization, drivers are able to complete the lap with a close to competitive lap time.$^\text{3}$
\paragraph*{Iteration - Adaptation rules and quantities} The general adaptation strategy seems to be similar for all drivers, where adaptation of the braking, i.e. brake points and peak brake pressure is particularly important.$^\text{3}$
By fine-tuning them, drivers manage to achieve better performances.$^\text{3}$
\par
\begin{figure}
	\centering
	\includegraphics[width=\columnwidth]{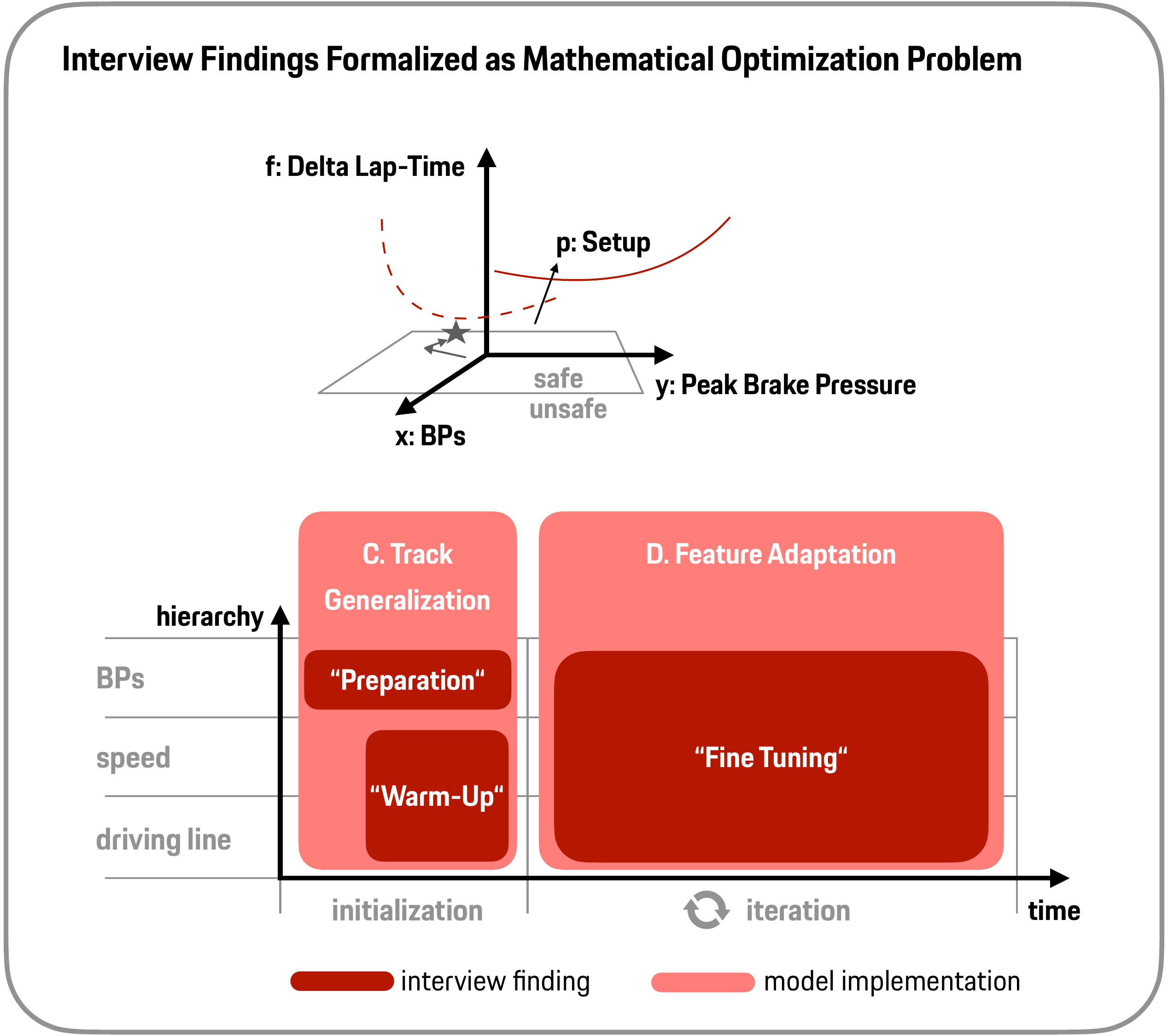}
	\caption{Top: Iterative adaptation process visualized as an optimization problem, initialized within a safe set by the drivers' experience. The delta lap-time is the objective function, parameterized by the vehicle setup. Variables include the brake points (BPs) and the peak brake pressure. Bottom: Three phases of driver adaptation to solve the above optimization problem, arranged in hierarchy-time plane. Dark color denotes findings from the expert interview and related work, whereas light color signifies how the respective findings are implemented in the adaptive model.}
	\label{fig:bigpicture}
\end{figure}

To summarize and simplify the problem, we set up the following qualitative model: Race drivers optimize delta lap-time as a function of brake points, peak brake pressure, and other variables as visualized in Figure \ref{fig:bigpicture}.
This function is parameterized through the vehicle setup.
To solve this problem, the brake points variable is initialized in the ``Preparation" phase in a safe region, i.e., such that the lap can be completed. Speed and driving line are initialized in hierarchical order during the ``Warm-Up" phase.
Afterward, drivers iteratively adapt and try out changes on all three hierarchical levels during ``Fine-Tuning".
Eventually, they arrive close to the optimizer shown as a star on the top of $\text{Figure \ref{fig:bigpicture}}$.
This point usually lies close to the boundary of the safe set, as the driver will be operating the vehicle at the limits of handling.
\subsection{ProMoD}
\label{sec:promod_original}
\begin{figure}
	\centering
	\includegraphics[width=0.8\columnwidth]{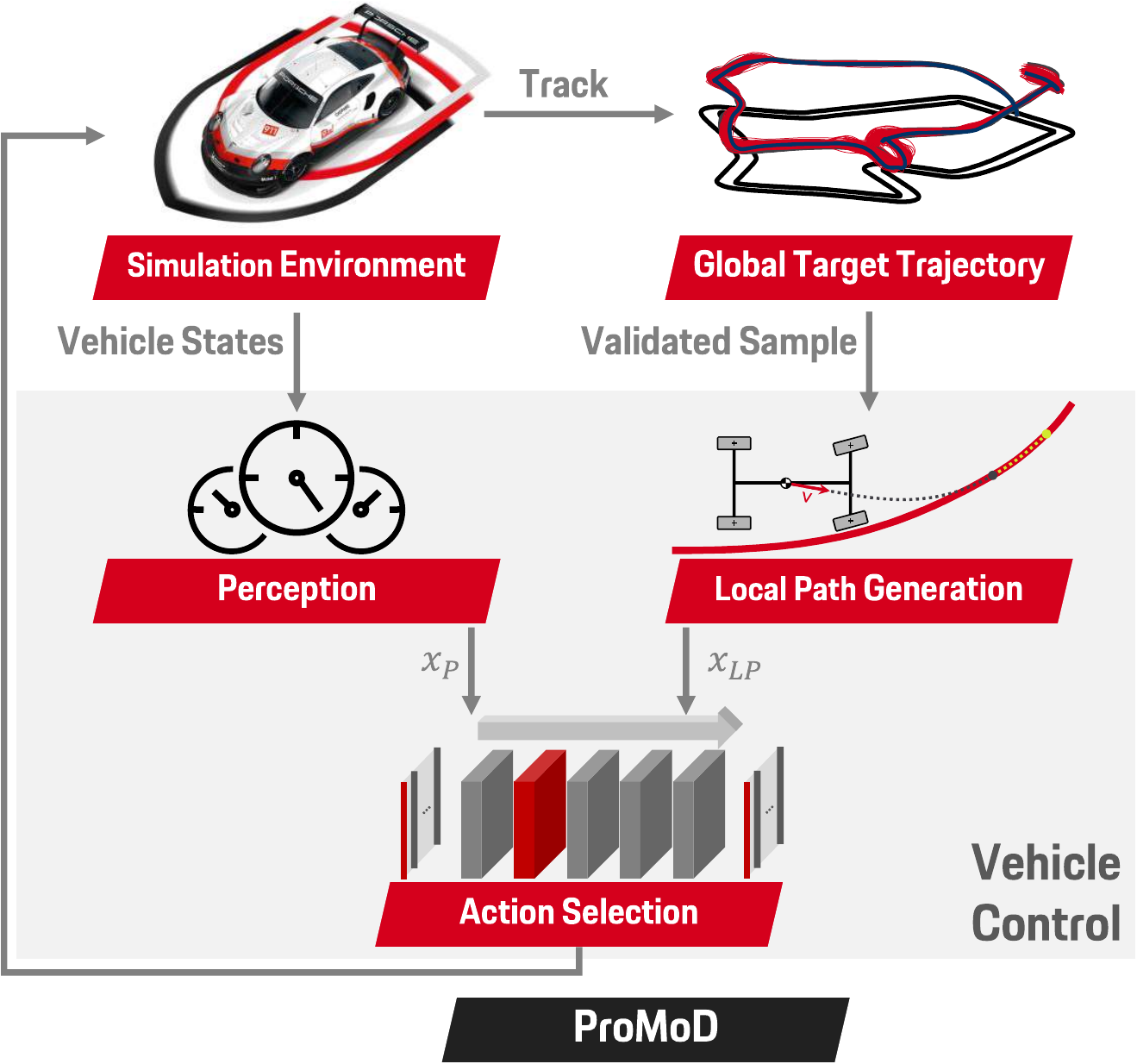}
	\caption{Original ProMoD framework to imitate human race drivers in a simulated environment with modules inspired from expert knowledge \cite{loeckel:2021:IMRD}:
	\textit{Global Target Trajectory} represents a distribution of potential target driving lines using ProMPs and relates to the driver's mental image of a driving corridor.
	\textit{Local Path Generation} and \textit{Perception} calculate a feature set based on the current situation on track and a sampled target driving line.
	The features are subsequently mapped to driver actions using a recurrent neural network in \textit{Action Selection}.
	Feeding back the predicted actions to the simulation environment closes the loop and facilitates offline simulation with a human-like driver model.
	}
	\label{fig:promod_framework}
\end{figure}

The recently presented ProMoD framework combines knowledge and ideas from both, race driver behavior and autonomous driving architecture.
It consists of multiple modules as visualized in Figure \ref{fig:promod_framework}, where each of these modules represents fundamental steps in the decision-making process of a human race driver. \cite{loeckel:2020:PFIH,loeckel:2021:IMRD}
In the following, the existing architecture is shortly summarized in order to provide a proper foundation for the subsequent development of the novel generalization and adaptation methods:
\paragraph*{Global Target Trajectory}
Every driver keeps a mental image of the whole race track in his head, knowing approximately where to brake, to turn in, and to accelerate again in each corner.
However, this imagined driving corridor is not precise, i.e. it incorporates some variance, and additionally changes over time with gathered experience.
Hence, we model the global target trajectory with a distribution over potential driving lines, which could be interpreted as a driving corridor, using ProMPs \cite{paraschos:2013:PMP, paraschos:2018:UPMP}.
For this purpose, both the spatial and the temporal information of every demonstrated driving line on a particular track should be projected to a lower-dimensional weight space.
We define a series of equally distributed Radial Basis Functions (RBFs)
\begin{equation} 
	b_{j}\left(s\right)=\exp \left(-\frac{\left(s-c_{j}\right)^{2}}{2 h}\right)	
	\label{eq:promp_bf},
\end{equation}
with function index $j \in \left\lbrace 1, 2, \ldots, N_{\mathrm{BF}} \right\rbrace$, track distance $s$, constant width $h$, and $c_j$ being the equally distributed centers of the functions.
All basis functions are assembled into the basis function matrix $\boldsymbol{\Phi}_s \in \mathbb{R}^{N_{\mathrm{s}} \times N_{\mathrm{BF}}}$, with $\boldsymbol{b}_{j}$ being the $j$th of $N_{\mathrm{BF}}$ columns, containing $b_{j}\left(s\right)$ evaluated at $N_{\mathrm{s}}$ equidistant points of the track distance.
Subsequently, $\boldsymbol{\Phi}_s$ is aggregated into
\begin{equation}
	\boldsymbol{\Psi}_s = \left[\begin{array}{cccc}
		\boldsymbol{\Phi}_{s, v_1} 		& \mathbf{0} 					& \cdots	& \mathbf{0} 	\\
		\mathbf{0} 						& \boldsymbol{\Phi}_{s, v_2} 	& \cdots	& \mathbf{0} 	\\
		\vdots							& \vdots						& \ddots	& \vdots		\\
		\mathbf{0} 						& \mathbf{0}					& \mathbf{0}& \boldsymbol{\Phi}_{s, v_{n}}
	\end{array}\right] \in \mathbb{R}^{nN_{\mathrm{s}} \times nN_{\mathrm{BF}}},
\end{equation}
for a total of $n$ variables that the trajectory consists of, with  $\boldsymbol{\Phi}_s = \boldsymbol{\Phi}_{s, v_1} = \cdots =  \boldsymbol{\Phi}_{s, v_{n}}$.
The weight vector
\begin{equation}
	\boldsymbol{w}_{i}=\left(\boldsymbol{\Psi}_s^{T} \boldsymbol{\Psi}_s+\epsilon \boldsymbol{I}\right)^{-1} \boldsymbol{\Psi}_s^{T} \boldsymbol{\tau}_{s, i}  \in \mathbb{R}^{nN_{\mathrm{BF}} \times 1}
	\label{eq:promp_wi}
\end{equation}
is derived using ridge regression for each demonstration trajectory $\boldsymbol{\tau}_{s, i}  \in \mathbb{R}^{nN_{\mathrm{s}} \times 1}$ and regularization factor $\epsilon$.
By fitting a Gaussian distribution $\mathcal{N}\left(\boldsymbol{\mu}_{\boldsymbol{w}}, \boldsymbol{\Sigma}_{\boldsymbol{w}}\right)$ over the $N$ demonstration weights with mean $\boldsymbol{\mu_{w}}$ and variance $\boldsymbol{\Sigma}_{\boldsymbol{w}}$
\begin{align}
	& \boldsymbol{\mu_{w}}=\frac{1}{N} \sum_{i=1}^{N} \boldsymbol{w}_{i}  \in \mathbb{R}^{nN_{\mathrm{BF}} \times 1}, \\
	& \boldsymbol{\Sigma}_{\boldsymbol{w}}=\frac{1}{N}\sum_{i=1}^{N}\left(\boldsymbol{w}_{i}-\boldsymbol{\mu}_{\boldsymbol{w}}\right)\left(\boldsymbol{w}_{i}-\boldsymbol{\mu}_{\boldsymbol{w}}\right)^{T}   \in \mathbb{R}^{nN_{\mathrm{BF}} \times nN_{\mathrm{BF}}}
	\label{eq:promp_dis}
\end{align}
we are able to describe the distribution of driving lines for a driver on a particular track efficiently.
Subsequently, an arbitrary number of new driving lines which are similar to all demonstrations can be generated by sampling a weight vector from this distribution 
$\boldsymbol{w}^{*} \sim \mathcal{N}\left(\boldsymbol{\mu}_{\boldsymbol{w}}, \boldsymbol{\Sigma}_{\boldsymbol{w}}\right)$
and using 
\begin{equation}
	\begin{aligned}
		\boldsymbol{\tau}^{*} = \boldsymbol{\Psi}_{s} \boldsymbol{w}^{*}
	\end{aligned}
	\label{eq:promp_ndis}
\end{equation}
to retrieve a new driving line in the original formulation which could be subsequently used for simulation.
\paragraph*{Local Path Generation}
For any situation on track, a human driver continuously plans the upcoming path a few seconds ahead.
We use this module to mimic the path planning by calculating constrained polynomials and multiple preview features\footnote{Examples are a predicted lateral offset or a predicted speed difference from the target driving line.
More details are given in \cite{loeckel:2021:IMRD}.} based on the current vehicle state and a sample from the previously constructed driving line distribution.
These local path features are denoted as $\mathbf{x}_\mathrm{{LP}}$.
\paragraph*{Perception}
In addition to the path planning features, each driver relies on additional information about their surroundings, such as visual information or experienced accelerations.
These perception features, which mostly relate to basic vehicle states, are gathered inside this module and are denoted as $\mathbf{x}_\mathrm{{P}}$.
\paragraph*{Action Selection}
The action selection process, i.e. the mapping from the current situation on track (as described by the feature set $\mathbf{x} = \left[\mathbf{x}_\mathrm{{LP}} ~ \mathbf{x}_\mathrm{{P}} \right]$) to human-like control actions $\mathbf{a}$, is learned using a recurrent neural network.
This neural network is trained on all available demonstration data for a particular driver, aiming to imitate its individual driving style and incorporating the dynamics of the action selection process.
\par
This architecture now allows to directly adapt the driver model on different levels in a transparent manner.
In contrast to end-to-end learning, a black-box behavior is avoided in order to increase interpretability.
In the following, we present methods to generalize and adapt this driver model in two different phases.
First, the `Track Generalization" is introduced in Section \ref{sec:trackgen} to address the ``Preparation"  and ``Warm-Up" steps identified from the interview (see Figure \ref{fig:bigpicture}).
Afterward, the iterative ``Fine-Tuning" is modeled by ``Feature Adaptation" in Section \ref{sec:featureadapt}.
The overall adaptation process is visualized in Figure \ref{fig:bigpicture}.
\subsection{Track Generalization: Generate Driving Line Distributions}
\label{sec:trackgen}
In order to generate first laps on a new, yet unknown track, it is required to learn a reasonable driving line distribution for the \textit{Global Target Trajectory} module.
All other modules of ProMoD are track-independent by definition and remain unmodified.
Hence, we aim to estimate a reasonable driving line distribution for this new track only based on the known track borders and on available experience from other tracks.
Inspired by the results from Section \ref{sec:expert_interview}, we propose the methodology described in Algorithm \ref{alg:driving_line}.
We utilize a novel ProMP description, conventional methods to fit driving lines based on geometric boundaries, and a method to estimate the variance of the driving line around the track based on experience from other tracks.
\begin{figure}
    \begin{minipage}{\columnwidth}
		\begin{algorithm}[H]
			\caption{Estimating a driving line distribution + sampling}
			\label{alg:driving_line}
			\begin{algorithmic}
				\State $\boldsymbol{\mu}_{\boldsymbol{w}}^{\kappa,dy},\boldsymbol{\Sigma}_{\boldsymbol{w}}^{\kappa,dy} \leftarrow$ \Call{BuildProMP}{$\mathcal{D}$}
				\State $x'\left(s\right),y'\left(s\right),\kappa'\left(s\right) \leftarrow$ \Call{BuildDrivingLine}{$\mathcal{B}_\mathrm{left}, \mathcal{B}_\mathrm{right}$}
				\State $\boldsymbol{\mu}_{\boldsymbol{w}}^{\kappa'} \leftarrow$ \Call{RidgeRegression}{$\kappa'\left(s\right)$}
				\State $\boldsymbol{\mu}_{\boldsymbol{w}}^{dy'} \leftarrow \mathbf{0}$
				\State $\boldsymbol{\Sigma}_{\boldsymbol{w}}^{dy'} \leftarrow$ \Call{EstimateVariance}{$\boldsymbol{\mu}_{\boldsymbol{w}}^{\kappa'}, \boldsymbol{\mu}_{\boldsymbol{w}}^{\kappa,dy}, \boldsymbol{\Sigma}_{\boldsymbol{w}}^{\kappa,dy}$}
				
				\State
				\For{$i \leftarrow 1, N_\mathrm{samples}$}
					\State $\boldsymbol{w}_i^{*dy} \sim \mathcal{N} \left(\boldsymbol{\mu}_{\boldsymbol{w}}^{dy'}, \boldsymbol{\Sigma}_{\boldsymbol{w}}^{dy'} \right)$
					\State $x_i^*\left(s\right),y_i^*\left(s\right) \leftarrow$ \Call{Reconstruct}{$x'\left(s\right),y'\left(s\right),\boldsymbol{w}_i^{*dy}$}			
					\State $\Delta t_i^*\left(s\right) \leftarrow$ \Call{EstimateSpeed}{$x_i^*\left(s\right),y_i^*\left(s\right),\mathcal{P}$}						
				\EndFor
			\end{algorithmic}
		\end{algorithm}	
    \end{minipage}
\end{figure}		
\paragraph*{ProMPs on Demonstration Data}
As drivers utilize their existing experience during familiarization with a new track, we are required to encode this knowledge in a reasonable way.
For this purpose, we use all available driving line data from all known tracks $\mathcal{D}$ and calculate ProMPs with a modified representation as driving line distributions for each track separately.
Hence, we take the time-based vehicle positions in the inertial reference frame for all laps on this track and map them to a curvilinear description
\begin{equation}
	x\left(t\right),~y\left(t\right) \mapsto dy\left(s\right),~\kappa\left(s\right)
\end{equation}
for each track.
Thereby, $dy$ represents the lateral deviation from a reference line and $\kappa$ the line curvature, both based on the reference line distance $s$.
While the information in $dy$ and $\kappa$ is partially redundant, both formulations are required for subsequent calculations.
By using equidistant samples from $dy$ and $\kappa$ and equidistantly spaced RBFs, it is now possible to project the driving line from each lap to a lower-dimensional weight space using ridge regression, resulting in weight vectors $\boldsymbol{w}^{dy}$ and $\boldsymbol{w}^{\kappa}$.
Assuming a Gaussian distribution, we retrieve mean weight vectors $\boldsymbol{\mu}_{\boldsymbol{w}}^{\kappa}$, $\boldsymbol{\mu}_{\boldsymbol{w}}^{dy}$ and variances $\boldsymbol{\Sigma}_{\boldsymbol{w}}^{\kappa}$, $\boldsymbol{\Sigma}_{\boldsymbol{w}}^{dy}$ to describe the distribution of all available driving lines on a particular track.
By iterating this process for all available tracks, we can aggregate all driving line information into $\boldsymbol{\mu}_{\boldsymbol{w}}^{\kappa,dy}$, $\boldsymbol{\Sigma}_{\boldsymbol{w}}^{\kappa,dy}$.
In the following, we estimate a driving line distribution for a unknown track by combining this variance information with a conventional path planning method.
\paragraph*{Generate Mean Driving Trajectory}
We start by estimating a mean driving trajectory which is only based on the known track boundaries $\mathcal{B}_\mathrm{left}$ and $\mathcal{B}_\mathrm{right}$.
As the generation of a reasonable and collision-free path around the track is required, we decide to use Elastic Bands \cite{quinlan:1993:EBCP,gehrig:2007:CAVF}.
While being computationally efficient and easy to interpret, this method showed to produce reasonable driving line estimates with sufficient accuracy.
The resulting trajectory is now assumed to be the reference and the mean driving line for the new track.
As it is initially represented in the inertial space $x'\left(s\right)$, $y'\left(s\right)$ we subsequently project it to the curviliniar space $\kappa'\left(s\right)$.
Similarly to the ProMP calculation on the available demonstration data, the curvature $\kappa'\left(s\right)$ is finally projected to the lower dimensional weight space and assumed to be the mean curvature $\boldsymbol{\mu}_{\boldsymbol{w}}^{\kappa'}$ with $\boldsymbol{\mu}_{\boldsymbol{w}}^{dy'} = \mathbf{0}$ by definition.
\paragraph*{Variance Estimation}
Using this mean trajectory and the existing corner information from other tracks, we estimate the variance with a sliding window approach.
For this purpose, we are moving along the estimated mean driving line's curvature $\kappa'\left(s\right)$ and compare the current situation, described by a sequence of curvatures, to all situations on all known tracks as encoded in $\boldsymbol{\mu}_{\boldsymbol{w}}^{\kappa,dy}$, $\boldsymbol{\Sigma}_{\boldsymbol{w}}^{\kappa,dy}$.
By finding the most similar corner measured by the absolute difference between curvatures, we are now able to iteratively build $\boldsymbol{\Sigma}_{\boldsymbol{w}}^{dy'}$, which describes the variance of driving lines on the new track.
\footnote{We use the curvature $\kappa$ to find similar corners since it naturally describes the corner shape.
The lateral deviation $dy$ is used for sampling, as it allows for a more robust reconstruction.}
\paragraph*{Sampling and Reconstruction}
Using the initially estimated mean driving line described by $x'\left(s\right)$, $y'\left(s\right)$, and the modified ProMP defined by mean $\boldsymbol{\mu}_{\boldsymbol{w}}^{dy'} = \mathbf{0}$ and covariance $\boldsymbol{\Sigma}_{\boldsymbol{w}}^{dy'}$ which describes the distribution of lateral deviations from this mean line, we are now able to sample an arbitrary number of new driving lines for the new track.
For this purpose we draw a sample weight $\boldsymbol{w}_i^{*dy} \sim \mathcal{N} \left(\boldsymbol{\mu}_{\boldsymbol{w}}^{dy'}, \boldsymbol{\Sigma}_{\boldsymbol{w}}^{dy'} \right)$ and retrieve the lateral deviation $dy_i^*\left(s\right)$ through reconstruction with $\boldsymbol{\Phi}_{s, dy} \boldsymbol{w}_i^{*dy}$.
Now, it is possible to construct the sampled driving trajectory in the Cartesian space using
\begin{align}
x_i^*\left(s\right) &= x'\left(s\right) - \mathrm{sin}\left(\phi_i^*\left(s\right) \right) dy_i^*\left(s\right)\\
y_i^*\left(s\right) &= y'\left(s\right) + \mathrm{cos}\left(\phi_i^*\left(s\right) \right) dy_i^*\left(s\right)
\end{align}
where $\phi_i^*$ corresponds to the heading angle of the vehicle, with $\phi_i^*=0$ when the vehicle drives purely into x-direction.
\paragraph*{Speed Profile}
In addition to the trajectory of the vehicle, ProMoD requires a speed profile for the \textit{Local Path Generation} module.
Since this velocity profile is dependent on the vehicle and its setup, and hard to estimate using the available demonstration data, we follow a more robust approach based on vehicle dynamics.
For each sampled vehicle trajectory $x_i^*\left(s\right), y_i^*\left(s\right)$, we utilize a conventional lap time estimation approach based on the vehicle performance envelope $\mathcal{P}$ to retrieve an approximate speed profile \cite{siegler:2002:LTSR, siegler:2000:LTSC}.
\paragraph*{Simulation}
The sampled driving lines with corresponding speed profiles can now be used to reconstruct the original ProMP formulation within the previously presented ProMoD framework.
An initialization with a reduced performance envelope $\mathcal{P}$ represents the ``Preparation"  phase on a new track and allows to safely simulate first laps.
By iteratively expanding $\mathcal{P}$ and simulating the resulting driving lines and speed profiles, ProMoD is able to cautiously approach the vehicle limitations, aiming to mimic the ``Warm-Up" phase.
The complete process facilitates simulations on a new track where no demonstration data exists, enhancing our driver modeling framework with track familiarization abilities to generate first fast laps.
However, as a human driver continuously optimizes its performance when being familiar with a track, ProMoD should also be adaptable and  learn from experience, as shown in Section \ref{sec:expert_interview}.
\subsection{Feature Adaptation}
\label{sec:featureadapt}
\begin{figure}
	\centering
	\includegraphics[width=\columnwidth]{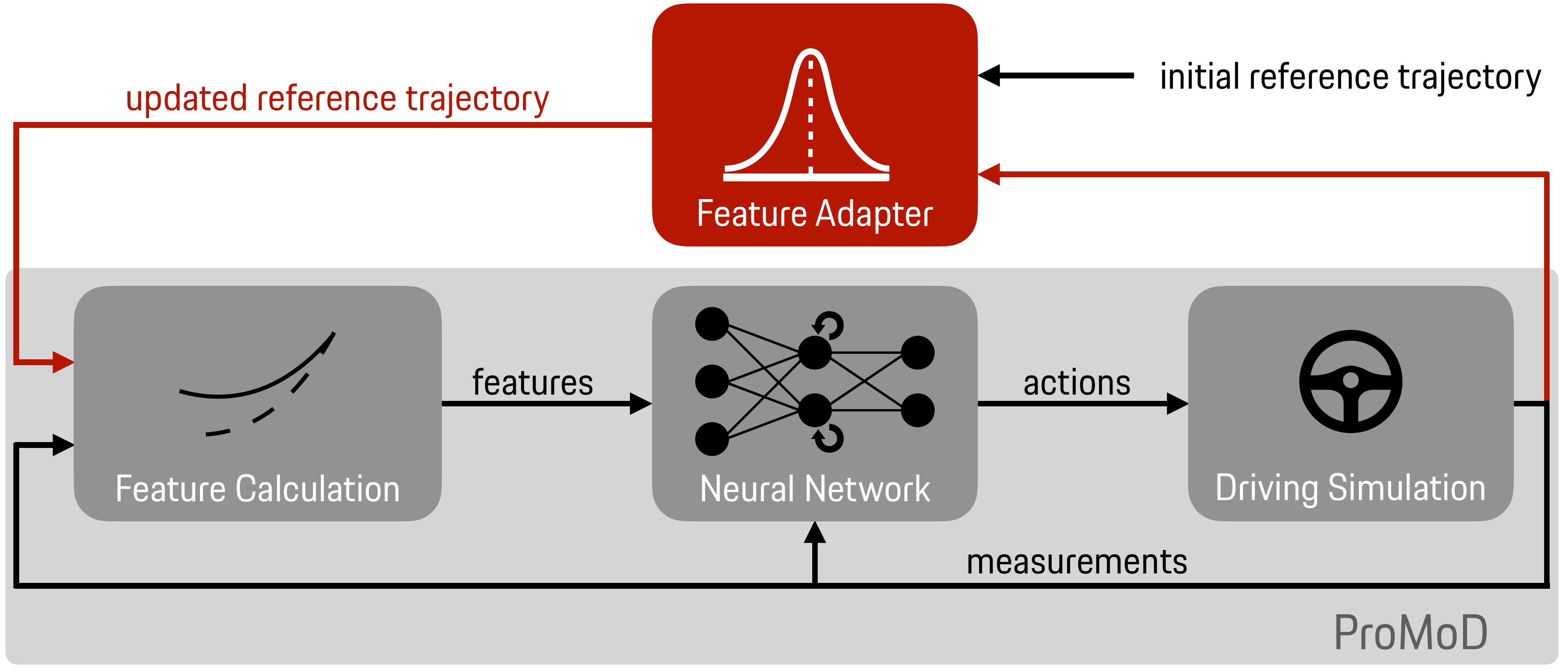}
	\caption{Structural design of feature adaptation:
The original ProMoD framework consists of three essential blocks: feature calculation, the neural network, and the simulation environment.
These elements remain unchanged during adaptation.
After each lap is simulated, the result is analyzed and the reference trajectory is adapted correspondingly. This new functionality is shown in red.}
	\label{fig:adaptation_feedback}
\end{figure}

Professional race drivers master the skill of continuously optimizing their performance by analyzing past laps and adapting their driving behavior accordingly.
With an additional feedback loop shown in Figure \ref{fig:adaptation_feedback}, ProMoD is enabled to mimic this learning process to a certain extent and achieve self-adapting behavior.
By only adapting the global target trajectory, which is used to compute local path planning features $\mathbf{x}_\mathrm{{LP}}$, the behavior of ProMoD can be influenced.
At the same time, ProMoD maintains its ability to imitate human drivers without yielding super-human performance as the action selection module remains unchanged.
In the following, we will use \textit{Conditioning} and \textit{Scaling} to effectively vary the trajectory while keeping it human-like:
\paragraph*{Conditioning} Recall that the ProMPs for the global target trajectory are represented by a Gaussian weight distribution  $p(\boldsymbol{w}) \sim \mathcal{N}\left(\boldsymbol{w} \mid \boldsymbol{\mu}_{\boldsymbol{w}}, \boldsymbol{\Sigma}_{\boldsymbol{w}}\right)$ with mean weight vector $\boldsymbol{\mu_w}$ and covariance matrix $\boldsymbol{\Sigma_w}$.
We are now able to alter this distribution by conditioning the prior distribution to a new observation $\boldsymbol{x}_{s'}^{*} = \{\boldsymbol{y}_{s'}^{*}, \boldsymbol{\Sigma}_{y}^{*}\}$ at a specific location $s = s'$, as presented in \cite{paraschos:2018:UPMP}.
Here, $\boldsymbol{y}_{s'}^{*} \in  \mathbb{R}^{n \times 1}$ is an algorithmically chosen target state of the vehicle position and velocity to be reached at distance $s'$, and variance $\boldsymbol{\Sigma}_{y}^{*} \in \mathbb{R}^{n \times n}$ is the confidence of this new observation.
The conditional distribution $p\left(\boldsymbol{w} \mid \boldsymbol{x}_{s'}^{*} \right)$ remains Gaussian with updated parameters
\begin{align}
	\boldsymbol{\mu}_{\boldsymbol{w}}^{[\mathrm{new}]} &=\boldsymbol{\mu}_{\boldsymbol{w}}+\boldsymbol{L}\left(\boldsymbol{y}_{s'}^{*}-\boldsymbol{\Psi}_{s}^{T} \boldsymbol{\mu}_{\boldsymbol{w}}\right), \\
	\boldsymbol{\Sigma}_{\boldsymbol{w}}^{[\mathrm{new}]} &=\boldsymbol{\Sigma}_{\boldsymbol{w}}-\boldsymbol{L} \boldsymbol{\Psi}_{s'}^{T} \boldsymbol{\Sigma}_{\boldsymbol{w}},
\end{align}
where
\begin{equation} \label{eq:L}
	\boldsymbol{L}=\boldsymbol{\Sigma}_{\boldsymbol{w}} \boldsymbol{\Psi}_{s'}\left(\boldsymbol{\Sigma}_{y}^{*}+\boldsymbol{\Psi}_{s'}^{T} \boldsymbol{\Sigma}_{\boldsymbol{w}} \boldsymbol{\Psi}_{s'}\right)^{-1}
\end{equation}
relates the variances of the prior distribution and the new observation with $\boldsymbol{\Psi}_{s'} \in \mathbb{R}^{nN_{\mathrm{BF}} \times n}$ representing the value of all basis functions at $s=s'$. \cite{paraschos:2018:UPMP}
\par
\begin{figure}
	\centering
	\begin{subfigure}[htb]{0.4\columnwidth}
		\includegraphics{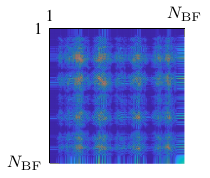}
		\caption{Orig. cov. $\boldsymbol{\Sigma}_{\Delta t \Delta t}$ }
		\label{fig:orig_var}
	\end{subfigure}
	\begin{subfigure}[htb]{0.4\columnwidth}
		\includegraphics{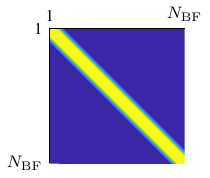}
		\caption{Factor matrix $\mathcal{F}_{k}$}
		\label{fig:factor_matrix}
	\end{subfigure}
	\begin{subfigure}[htb]{0.6\columnwidth}
		\includegraphics{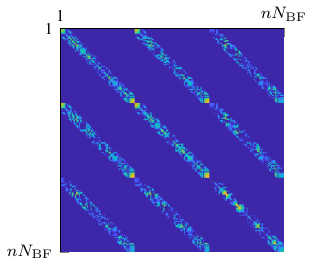}
		\caption{Pre-processed cov. $\boldsymbol{\Sigma}_{\boldsymbol{w}}^{\text{masked}}$}
		\label{fig:filtered_var}
	\end{subfigure}
	\caption{Masking the covariance matrix:
(a) Part of the covariance matrix for a single variable \text{($\boldsymbol{\Sigma}_{\Delta t \Delta t}  \in \mathbb{R}^{N_{\mathrm{BF}} \times N_{\mathrm{BF}}}$)}, where brighter colors indicate higher covariances.
Far-off-diagonal correlations in the data potentially result from different vehicle setups in the demonstration data but are difficult to consider during conditioning.
(b) Factor matrix for a single variable, where the elements on the diagonal are one, and off-diagonal entries are fading out to zero using bandwidth $k$.
Here, $k$ is selected such that distant and non-consecutive turns can not mutually influence each other.
(c) Resulting matrix $\boldsymbol{\Sigma}_{\boldsymbol{w}}^{\text{mask}}  \in \mathbb{R}^{n N_{\mathrm{BF}} \times n N_{\mathrm{BF}}}$ for three variables after masking, filtering out correlations over larger distances.}
	\label{fig:adapt_variance}
\end{figure}

This procedure allows to move brake points or to shift apexes\footnote{An apex is defined as the closest point to the inner side of a corner, typically coinciding with the locally maximal curvature of the driving line.} by conditioning the prior distribution utilizing a set of rules derived from Section \ref{sec:expert_interview}.
However, when using the prior variance without further consideration, conditioning at a specific turn potentially affects distant turns due to non-zero covariances in the data, as shown for $\boldsymbol{\Sigma}_{\Delta t \Delta t}$ in Figure \ref{fig:adapt_variance} (a).
As such a large effect across multiple turns would not be considered human-like, we aim to reduce it by masking the original matrix using a factor matrix $\mathcal{F}_{k} \in \mathbb{R}^{N_{BF} \times N_{BF}}$ shown in Figure \ref{fig:adapt_variance} (b).
By multiplying $\mathcal{F}_{k}$ element-wise with each submatrix of $\boldsymbol{\Sigma}_{\boldsymbol{w}}$, we retrieve a masked matrix for conditioning
\begin{equation}
	\boldsymbol{\Sigma}_{\boldsymbol{w}}^{\text{masked}} = \left[\begin{array}{ccc}
		\mathcal{F}_{k}\circ\boldsymbol{\Sigma}_{xx} & \mathcal{F}_{k}\circ\boldsymbol{\Sigma}_{xy} & \mathcal{F}_{k}\circ\boldsymbol{\Sigma}_{x\Delta t} \\
		\mathcal{F}_{k}\circ\boldsymbol{\Sigma}_{yx} & \mathcal{F}_{k}\circ\boldsymbol{\Sigma}_{yy} & \mathcal{F}_{k}\circ\boldsymbol{\Sigma}_{y\Delta t} \\
		\mathcal{F}_{k}\circ\boldsymbol{\Sigma}_{\Delta t x} & \mathcal{F}_{k}\circ\boldsymbol{\Sigma}_{\Delta t y} & \mathcal{F}_{k}\circ\boldsymbol{\Sigma}_{\Delta t \Delta t} \\
	\end{array}\right]
	\label{eq:variance_matrix_mask}
\end{equation}
which effectively lowers the influence on distant regions as shown in Figure \ref{fig:adapt_variance} (c).
This matrix could now be used for effective local \textit{Conditioning}.
\paragraph*{Scaling} In order to fully utilize the vehicle's potential on straights the speed profile can be adapted to influence the throttle actuation and braking behavior of ProMoD.
Since the neural network works to some extent like a trajectory-tracking controller, whose goal is to keep the control error between the reference speed and the actual speed as small as possible, its output signals tend to fluctuate during intervals of full throttle.
Therefore, if the actual velocity is larger than the reference velocity, ProMoD tends to accelerate less, even if the virtual driver is on a straight line and is expected to drive as fast as possible.
This problem can be effectively solved by smoothly scaling the reference speed on long straights.
\paragraph*{Adaptation Process} 
The complete adaptation process, shown in Algorithm \ref{alg:adaptation}, is inspired by the insights from Section \ref{sec:expert_interview} and uses both introduced methods, \textit{Conditioning} and \textit{Scaling}, to continuously adapt ProMoD based on gathered experience.
While the original version of ProMoD simply samples and simulates multiple driving lines from a single distribution, we will now use the variance information for a continuous and targeted adaptation.
\begin{figure}
    \begin{minipage}{\columnwidth}
		\begin{algorithm}[H]
			\caption{Adaptation Process}
			\label{alg:adaptation}
			\begin{algorithmic}
				\State \textbf{Input}: $\boldsymbol{\mu}^0_{\boldsymbol{w}}, \boldsymbol{\hat{\Sigma}}^0_{\boldsymbol{w}}$, envelope
				\State $\boldsymbol{\Sigma}^0_{\boldsymbol{w}} \gets $\Call{processVariance}{$\boldsymbol{\hat{\Sigma}}^0_{\boldsymbol{w}}$} 		
				\State $\boldsymbol{\tau}^{ref}_0 \gets$ \Call{calcMeanTrajectory}{$\boldsymbol{\mu}^0_{\boldsymbol{w}}$}
				\State $\boldsymbol{\mathcal{I}}^{track} \gets$ \Call{analyseTrack}{$\boldsymbol{\tau}^{ref}_0$} 
				\For {$i = 0, 1, 2, ...$}
				\State $\boldsymbol{\tau}_i \gets$ \Call{simulate}{$\boldsymbol{\tau}^{ref}_i$}
				\State $\boldsymbol{y}_{s}^{*} = \text{\O}$
				\If {not \Call{isCompleted}{$\boldsymbol{\tau}_i$}} 
				\State $\boldsymbol{y}_{s}^{*} \gets \boldsymbol{y}_{s}^{*} \cup$ \Call{analyseDL}{$\boldsymbol{\tau}_i, \boldsymbol{\mathcal{I}}^{track}, \text{envelope}$}
				\If {$\boldsymbol{y}_{s}^{*} == \text{\O}$ or $\textproc{slipCheck}(\boldsymbol{\tau}_i, \boldsymbol{\mathcal{I}}^{track})$}
				\State $\boldsymbol{y}_{s}^{*} \gets \boldsymbol{y}_{s}^{*} \cup$ \Call{adaptSpeed}{$\boldsymbol{\tau}_i, \boldsymbol{\mathcal{I}}^{track}$}
				\EndIf
				\Else 
				\State  $\boldsymbol{y}_{s}^{*} \gets \boldsymbol{y}_{s}^{*} \cup$ \Call{checkInEnvelope}{$\boldsymbol{\tau}_i, \text{envelope}$} 	  
				\EndIf
				\State $\boldsymbol{\mu}^{i+1, 0}_{\boldsymbol{w}}, \boldsymbol{\Sigma}^{i+1, 0}_{\boldsymbol{w}} = \boldsymbol{\mu}^{i+1}_{\boldsymbol{w}}, \boldsymbol{\Sigma}^{i+1}_{\boldsymbol{w}}$
				\For {$j = 1, 2, ...,$ number of items in $\boldsymbol{y}_{s}^{*}$} 
				\State $\boldsymbol{\mu}^{i+1, j}_{\boldsymbol{w}}, \boldsymbol{\Sigma}^{i+1, j}_{\boldsymbol{w}} \gets$ \Call{cond}{$\boldsymbol{\mu}^{i+1, j-1}_{\boldsymbol{w}}, \boldsymbol{\Sigma}^{i+1, j-1}_{\boldsymbol{w}}, \boldsymbol{y}_{s,j}^{*}$}	
				\EndFor
				\\
				\State $\boldsymbol{\hat{\tau}}^{ref}_{i+1} \gets$ \Call{calcMeanTrajectory}{$\boldsymbol{\mu}^{i+1}_{\boldsymbol{w}}$}	
				
				\If {\Call{isCompleted}{$\boldsymbol{\tau}_i$}}
				\State $\boldsymbol{\tau}^{ref}_{i+1} \gets$ \Call{speedScaling}{$\boldsymbol{\hat{\tau}}^{ref}_{i+1}, \boldsymbol{\tau}_i, \boldsymbol{\mathcal{I}}^{track}$}
				\Else
				\State $\boldsymbol{\tau}^{ref}_{i+1} \gets \boldsymbol{\hat{\tau}}^{ref}_{i+1}$ 
				\EndIf                                                                                                                                                                                          
				\EndFor
			\end{algorithmic}
		\end{algorithm}	
	\end{minipage}
\end{figure}	
After simulating a lap, an initial check is done whether the lap was completed successfully.
If the simulation ended before completing a full lap, the situation where the vehicle left the track is analyzed and the ProMP is conditioned using two sub-procedures:
\begin{itemize}
	\item Driving-line check and adaptation:
	As illustrated in Section \ref{sec:expert_interview}, the turn-in is the most important phase during cornering.
	Hence, the driving line is compared to the permissible driving corridor, represented by track borders or by the envelope of all demonstrations from the human drivers, and the largest deviation before the apex is found.
	Then, a new observation $\boldsymbol{y}_{s'}^{*}$ is added for \textit{Conditioning} at this position, shifting the driving line distribution towards the permissible area.
	\item Velocity adaptation:
	If no valid adaptation is found or extreme tire slip occurs, the target velocity will be reduced.	
\end{itemize}
In practice, ProMoD can eventually complete each critical corner when the target speed is low enough.
Subsequently, the completed laps can be further adapted to improve the lap time and to keep the driving line in the envelope by:
\begin{itemize}
	\item Checking and reducing smaller deviations from the permissible driving corridor:
	Just like in the real race, ProMoD sometimes slightly exceeds the theoretically allowed driving corridor but still manages to complete the lap.
	These situations are checked and additional control points are introduced for \textit{Conditioning}.
	\item Checking acceleration intervals and \textit{Scaling} of the speed:
	As discussed before, ProMoD partially does not utilize the full vehicle potential during acceleration phases on straight lines.
	Hence, speed scaling is used to further increase the performance on already completed laps.
\end{itemize}
\par
By introducing this process, we are able to encourage ProMoD to learn from the experience from previous laps, to correct mistakes, and to increase performance.
	\section{Evaluation}
In this work, we use data of professional race drivers generated with the HDiL simulator shown in Figure \ref{fig:simulator} to train and evaluate our driver modeling approach.
All rollouts of our driver model are simulated using the same in-house developed vehicle model of a high-performance race car, guaranteeing realistic vehicle dynamics and facilitating comparability to the human demonstrations.
The task of driving the simulated race car is highly challenging as the car only uses a \textit{Traction Control} as driver assistance system.
In order to safeguard intellectual property, all plots in this paper are shown normalized.
Furthermore, all driver names are anonymized.
\subsection{Track Generalization}
\begin{figure}
	\centering
	\includegraphics[width=1.0\columnwidth]{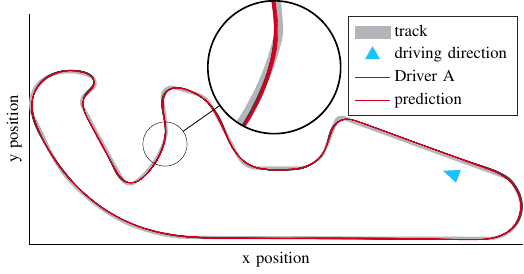}
	\caption{Prediction of the global target trajectory distribution on AGN:
	We compare 25 samples from the real driver (dark grey) to 25 samples from the predicted target trajectory distribution (red) derived with the driving line generalization method.
	The light grey area visualizes the track and the bright blue triangle indicates the position of the start/finish line and driving direction.
	Here, the method is able to approximately reproduce the expected driving lines with some partial deviations.
}
	\label{fig:pred_var_AGN}
\end{figure}

\begin{figure*}
	\begin{subfigure}{\textwidth}
		\centering
		\includegraphics[width=\textwidth]{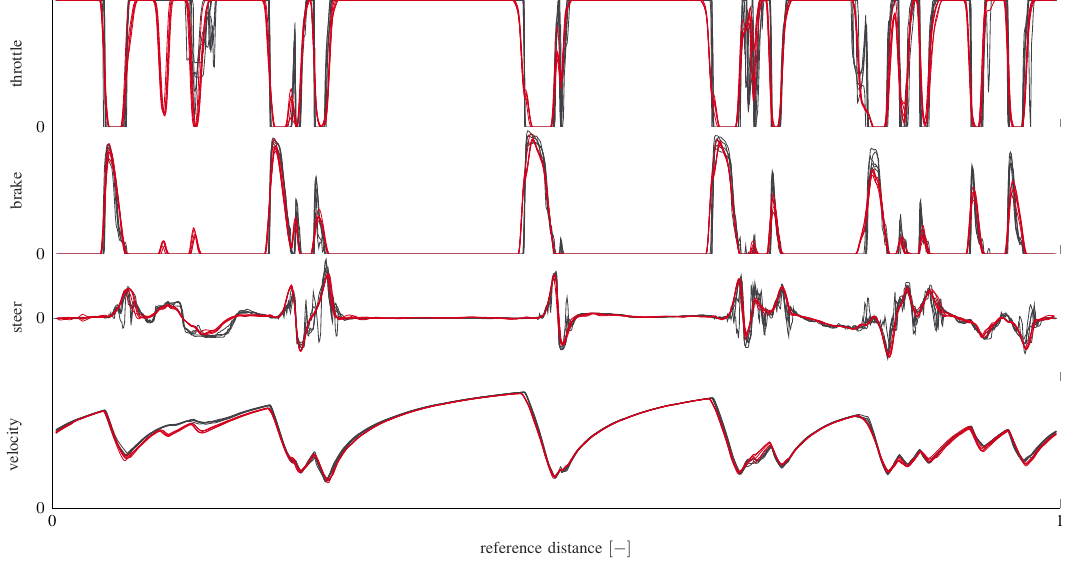}
		\caption{driver actions}
	\end{subfigure}\\
	\begin{subfigure}{\textwidth}
		\centering
		\includegraphics[width=\textwidth]{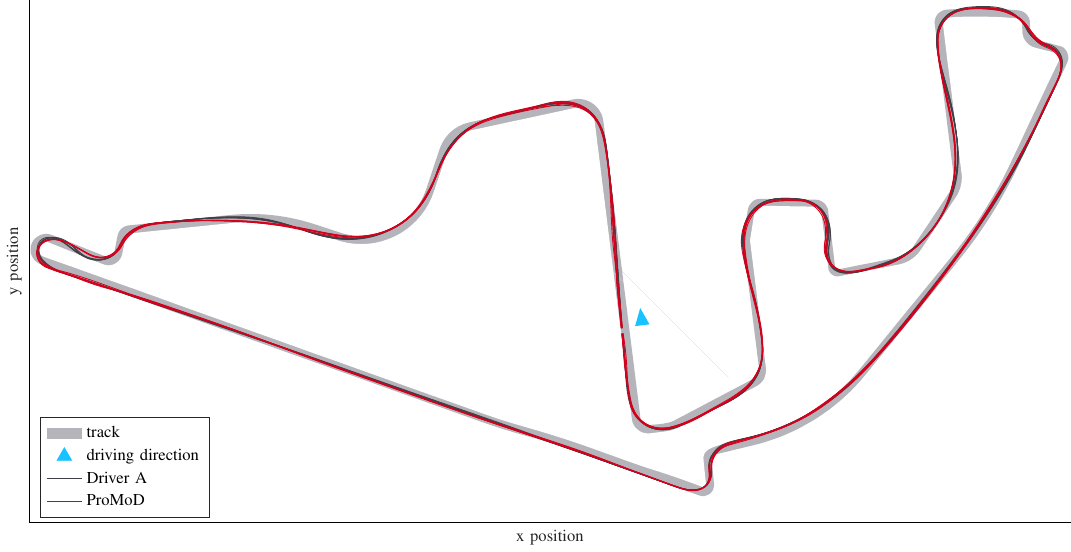}
		\caption{driving lines}
	\end{subfigure}	
	\caption{Track Generalization results on ABD:
	We compare five laps from the human driver (dark grey) to five laps from the track generalized ProMoD framework (red) on the identical vehicle setup.
	(a) shows a comparison of the driver actions and the resulting speed profiles over the normalized track reference distance.
	Here, ProMoD is able to approximately reproduce the throttle, braking, and steering activity of the real driver considering the braking points, actuation speeds, and amplitudes.
	The velocity profile shows some small deviations after the first corner where ProMoD does not fully utilize the vehicle potential due to a slightly too conservative speed profile estimation in this region.
	(b) visualizes the resulting simulated driving lines around the track (light grey).
	The position of the start/finish line and the driving direction is indicated by the bright blue triangle.
	Here, ProMoD approximately follows the demonstrations of the human driver despite they were not used for training.
	Some deviations are present at particularly challenging locations (e.g. the hairpin corner on the left side), which, however, do not prevent ProMoD from finishing the lap with a reasonable performance.
	These deviations may be reduced using adaptation methods to learn from the gathered experience on the track.}
	\label{fig:ABD}	
\end{figure*}

We evaluate the presented track generalization method of our ProMoD framework on two different race tracks, Motorland Arag\'{o}n (AGN) and the Yas Marina Circuit in Abu Dhabi (ABD), and leave out the correspondingly available demonstration data in the training procedure of our driver model.
We start by comparing the predicted driving line distribution to the corresponding driving lines of the human driver on AGN in Figure \ref{fig:pred_var_AGN}.
It is visible that the generated driving line distribution, despite not being completely driver-specific and showing some small variations, approximately fits the driven paths of the real driver.
When using these driving line distributions for simulation on unknown tracks, ProMoD is capable of completing full laps on the respective race track, as visualized in Figure \ref{fig:ABD} for ABD.
\par
\begin{figure}
	\centering
	\includegraphics[width=\columnwidth]{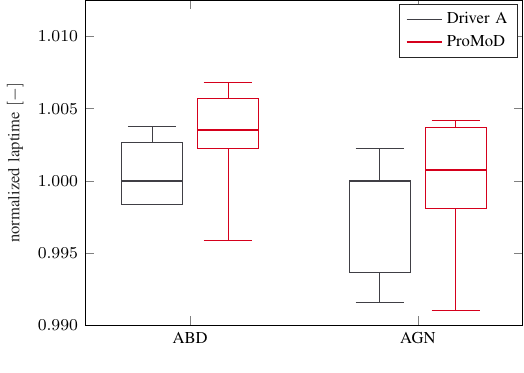}
	\caption{Lap time comparison for track generalization on race tracks ABD and AGN:
	Times are normalized to the median demonstration lap time of the corresponding track.
	The whiskers correspond to the minimum/maximum values, boxes indicate the upper/lower quartiles, and the thick central line shows the median value.
	Here, ProMoD is able to finish complete laps on the unknown race tracks being less than $0.5\%$ slower than the human driver in the median, but partially already achieving competitive times on particularly fast laps.
	The slightly slower median lap time might be a result of a yet non-optimal speed profile or driving line distribution.
	}
	\label{fig:dlg_laptimes}
\end{figure}

For AGN, the track generalization method achieves comparable results considering the similarities of the resulting driving line and driver action distributions with the human driver.
Furthermore, we compare the performances of ProMoD and the real driver on both tracks with equal vehicle setups.
Figure \ref{fig:dlg_laptimes} visualizes the resulting lap time distributions in a box plot, respectively normalized to the median lap time of the human driver on each track.
Here, ProMoD is able to achieve lap times close to the real driver, with a slightly increased median due to small deviations in the expected speed profiles.
\subsection{Feature Adaptation}
\begin{figure}
	\begin{subfigure}[tb]{\columnwidth}
		\includegraphics{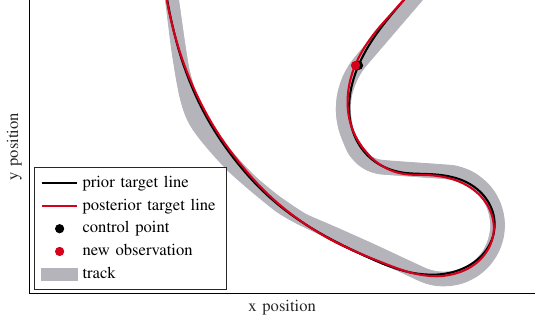}
		\caption{adaptation of the target line}
	\end{subfigure}
	\begin{subfigure}[tb]{\columnwidth}
		\includegraphics{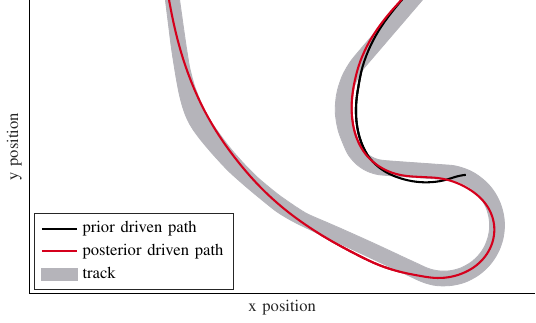}
		\caption{resulting prior and posterior driving lines}
	\end{subfigure}
	\caption{Adaptation of the target line for T6/7 on SVT and the resulting driven paths:
	(a) visualizes the prior (black) and posterior (red) target lines.
	The posterior target line is pulled outwards before the first apex using a control point at corner entry, as ProMod initially exceeds the vehicle potential and left the track.
	(b) shows the resulting driven paths from ProMoD within the complex simulation environment.
	After simultaneous adaptation of the target line and the velocity profile, ProMoD is able to successfully pass this turn without leaving the track.}
	\label{fig:adapt_conditioning_t06_xy}
\end{figure}

\begin{figure}
	\centering
	\includegraphics[width=\columnwidth]{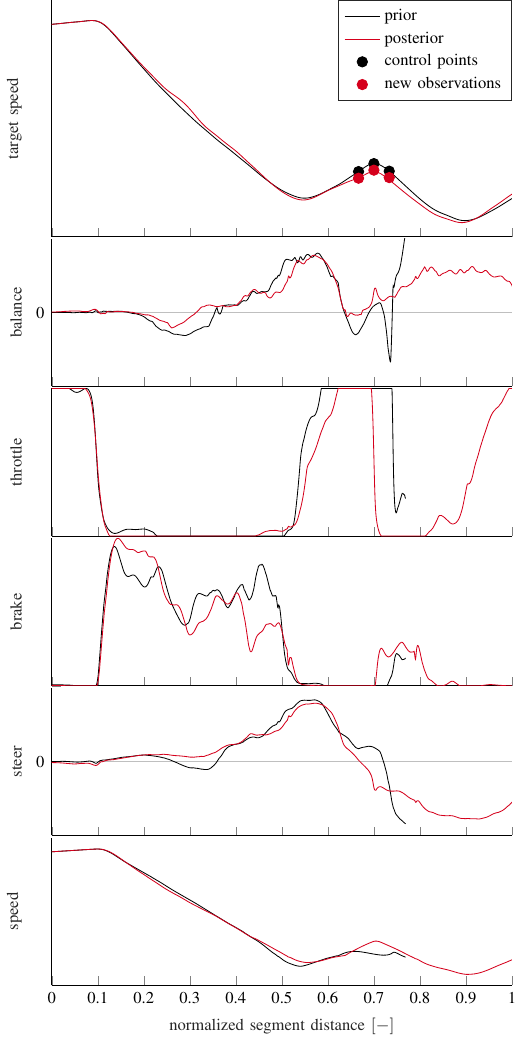}
	\caption{Adaptation of the target speed profile for T6/7 on SVT:
	This figure visualizes the target speed, as well as resulting vehicle states and driver actions over the normalized segment distance before and after adaptation.
	By using three control points the method is able to adapt the target speed profile effectively while preserving its general shape.
	The balance plot indicates the current driving state:
	Positive values relate to understeering, i.e. situations where the vehicle "plows" and turns less than a neutral steering car.
	Contrarily, negative values indicate oversteering situations, i.e. the limits of the rear axle are exceeded and the vehicle might start to spin if the driver does not counter steer.
	Before adaptation and at normalized segment distance 0.25, the vehicle gets in such an oversteering situation, but ProMoD is able to counter steer and recover the vehicle at the cost of losing speed.
	However, at distance 0.65 ProMoD largely exceeds the vehicle potential, resulting in a slide over both axles which forces the vehicle off the track (see Figure \ref{fig:adapt_conditioning_t06_xy} (b)).
	After the adaptation of the speed profile and the driving line, ProMoD is able to keep the vehicle on the track without exceeding its potential, considerably reducing critical situations.
	Here, ProMoD uses an increased braking force during the first turn in, later acceleration, and earlier throttle lift and braking for the following turn.
	Those changes contribute to the success of the adaptation and ProMoD passes the turn after two iterations.}
	\label{fig:adapt_conditioning_t06_v}
\end{figure}

\begin{figure}
	\centering
	\includegraphics{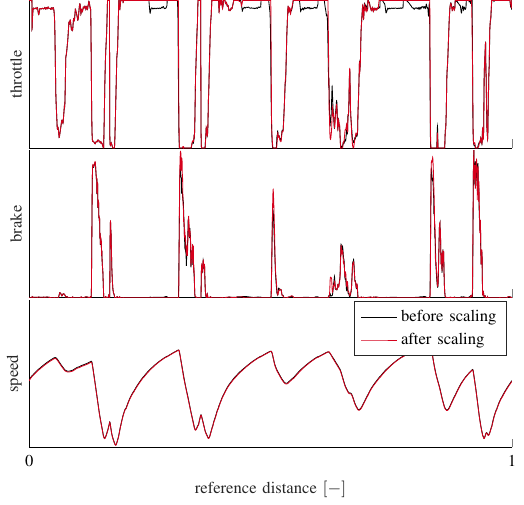}
	\caption{Effect of speed scaling on straight lines:
	After scaling, ProMoD effectively utilizes the longitudinal potential of the vehicle and uses full throttle on most straights.
	For intervals where ProMoD fails in subsequent turns due to the increased speed, scaling is prevented.}
	\label{fig:adapt_scaling}
\end{figure}

\begin{figure}
		\includegraphics{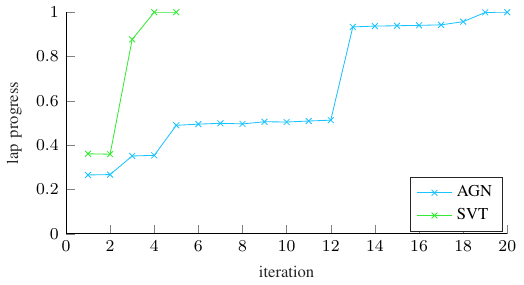}
	\caption{The adaptation progress on ProMoD on AGN and SVT:
	For both tracks, ProMoD succeeds in completing a previously unfinished lap within 20 iterations.
	Subsequent iterations can be used to further increase performance.}
	\label{fig:fa_svt_agn}
\end{figure}

The feature adaptation process is tested on two different tracks, the Silverstone Circuit (SVT) and Motorland Arag\'{o}n (AGN).
We start with an evaluation of the local effects of \textit{Conditioning} and \textit{Scaling} by showing the executed adaptations, the resulting changes in the driven path, and the selected actions of the driver model.
Subsequently, we are going to test the complete adaptation process on both tracks, showing that the method is able to increase the covered distance for previously unfinished laps, and to increase performance by reducing lap time without losing control.
\paragraph*{Local Effect - Adaptation}
The local effects of adaptation are presented in Figure \ref{fig:adapt_conditioning_t06_xy} and Figure \ref{fig:adapt_conditioning_t06_v}, visualizing adaptations of the driving line and the speed profile, as well as the resulting action signals and driven paths.
Here, ProMoD fails initially at Turn (T) 6/7 of the SVT racetrack due to considerably exceeding the vehicle potential as shown in Figure \ref{fig:adapt_conditioning_t06_xy} (b).
In order to adapt the speed profile effectively, three control points are used to set the lower peak speed value, resulting in earlier braking and consequently helping to avoid the mistake and pass the turn.
At the same time, with the purpose of reducing the curvature at corner entry, the driving line is pulled outwards around fifty meters before the first apex as shown in Figure \ref{fig:adapt_conditioning_t06_xy} (a).
After two iterations of simultaneously adapting both, the speed profile and the driving line, ProMoD succeeds in passing this turn.
\paragraph*{Local Effect - Scaling}
Scaling is particularly useful on straights, if ProMoD initially does not fully utilize the vehicle potential due to a modified vehicle setup and a too conservative prior target speed definition.
Its effect becomes apparent when observing the accelerator actuation signal.
With a higher reference speed, the model tends to utilize full throttle more often on long straight lines, as shown in Figure \ref{fig:adapt_scaling}. 
As a consequence, the fluctuation of the throttle signal on those intervals are effectively eliminated, and the lap time is improved by about $0.2 $ seconds. 
\paragraph*{Adaptation Process}
The developed adaptation process for ProMoD has been successfully tested on SVT and AGN racetracks as visualized in Figure \ref{fig:fa_svt_agn}.
While it only requires 4 iterations to complete SVT, ProMoD needs more iterations for AGN since it fails at more locations.
On both tracks, ProMoD succeeds in completing a lap after less than 20 iterations, with at most five iterations for a problematic turn.
\FloatBarrier
	\section{Conclusion}
In this paper, we present new insights into the general adaptation behavior and the learning processes of professional race drivers and derive new methods to extend ProMoD, an advanced modeling method for driver behavior.
With the purpose of understanding driver behavior in general and identifying the most important adaptation processes, this work starts with key insights from related work and an expert interview with a professional race engineer.
Based on the hereby acquired knowledge, we develop a novel method that can estimate human-like driving line distributions for unknown tracks.
These distributions could be used to simulate complete laps with almost competitive performances and human-like driver control inputs in a professional motorsport driving simulator.
Subsequently, we present a feature adaptation method that allows ProMoD to learn from the gathered experience of previous laps.
Using different experiments, we demonstrate the ability to continuously learn from mistakes and to improve driving performance.
\par
This work contributes to the modeling and a better understanding of driver behavior, paving the way for advanced full-vehicle simulations with consideration of the human driver and potentially future autonomous racing.
\par
Due to its modular architecture, ProMoD could be extended in various ways in future research.
Besides additional methods for feature adaptation and optimization, the neural network of the action selection module could be adapted to learn from experience using reinforcement learning techniques, or real track data may be used to provide more demonstration data.
Furthermore, human-like qualitative feedback, which is based on encountered problems during driving, could help to further support the vehicle development process.
In addition, our driver model may be extended to a multi-agent environment with opponents on the race track, facilitating a more accurate prediction of true racing performance and potentially optimization of complete racing strategies.
Finally, ProMoD might be applied to other similar use cases, with the target of modeling human behavior in dynamic environments with small stability margins.	
	\bibliography{ms}
	\section{Appendix - Expert Interview}
\label{app:expertinterview}
\textbf{Is there a universal adaption rule that applies to all drivers and tracks?} 
\\
\textit{Indeed, it turns out that adaption strategies are very similar across different drivers, tracks, and vehicles, in spite of the individual driving behavior, the various layouts of the tracks and the continuously modified vehicle setups.
The driver's main goal is to `brake as late as possible, and accelerate as early as possible'.
The resulting driving line, the turn-in, and the on-throttle behavior are seen as a consequence of pursuing that goal.}
\par
\textbf{How do drivers drive their first laps on a new track?}
\\
\textit{When faced with a new track, what a driver would do can be divided into three phases: preparation, warm-up, and the subsequent fine tuning.}
\begin{itemize}
	\item \textit{Preparation.
	Drivers come to a new track with a memorized `database of corner information', collected from their prior experience, simulator sessions, statistical data, etc.
	First, drivers characterize each new corner by comparing it with those in their memory, and assemble a first guess of the driving line.
	Since every corner is unique, this first guess is usually a rough approximation.
	At this point, it is helpful to consult other drivers to improve the initial guess.
	Finally, they set brake points, utilizing signs in the environment such as brake markers.
	Having concretized all prior information and exchanged opinions with fellow drivers of specific positions for hitting the brake pedal, the drivers start their first laps on a new track.}
	\item \textit{Warm-up.
	Race drivers are particularly talented in assessing risk. 
	They usually start off with a slow and safe speed profile, which they adapt from lap to lap to higher velocities.
	This process can take very few iterations.
	For example, one driver managed to reach a competitive lap time on the Le Mans circuit surprisingly after only five laps.}
	\item \textit{Fine Tuning.
	 After warming up, drivers are able to complete the lap with a close to competitive lap time, which they then try to improve incrementally.
	 Usually, drivers do not reach a global optimum, but are aware of how to improve.
	 High- and changing-speed corners are the most difficult ones, where spinning should be prevented, as it is extremely difficult to control.}
\end{itemize}
\par
\textbf{Which quantities do race drivers adapt and how? Do they pay attention to specific metrics?}
\\
\textit{Although the goal of improving lap time is sound and clear, the real optimization process is indeed very complicated, and many factors have to be taken into consideration.
The following three aspects are most critical during optimization:}
\begin{itemize}	
	\item \textit{Delta lap time.
	The adaption behavior of race drivers is result-oriented.
	They are not paying much attention to the exact speed values at local points around the track, but rather to the lap time difference to the previous or best lap.}
	The association with the optimization problem is visualized on the top of Figure \ref{fig:bigpicture}.
	\item \textit{Brake point.
Hitting the brake is where the corner starts.
It is the most crucial tuning knob, not only because it influences the speed profile, but also since it is the source of any issues arising throughout the following corner.
I.e., all issues should be traced back to the brake point, and cannot be locally analyzed.}
	\item \textit{Peak brake pressure.
The driver attempts to predict the future state of the car when making decisions.
In the presence of slip, however, uncertainty about the vehicle state is introduced, eventually leading to wrong predictions by the driver.
Therefore, slip management is crucial during cornering, with the maixmum brake pressure helping to anticipate imminent slip.}
\end{itemize}
\par
\textbf{How do race drivers behave when the vehicle setup is modified? Will they pre-adapt their strategy according to the setup?}
\\
\textit{It is extremely complicated to analyze the car and the behavior of the driver simultaneously.
Therefore, when new vehicle setups are tested, the drivers do not and are not expected to have much  idea of what has been adapted on the car.
Sometimes,  race engineers would do blind-tests in order to isolate the influences of the modified setups from those of the drivers.}
\end{document}